\documentclass[letterpaper]{article} 
\usepackage{aaai23}  
\usepackage{times}  
\usepackage{helvet}  
\usepackage{courier}  
\usepackage[hyphens]{url}  
\usepackage{graphicx} 
\usepackage{natbib}  
\usepackage{caption} 
\usepackage{subfig}

\usepackage{algorithm}
\usepackage{algorithmic}

\usepackage{amsfonts}
\usepackage{newfloat}
\usepackage{listings}
\DeclareCaptionStyle{ruled}{labelfont=normalfont,labelsep=colon,strut=off} 
\floatstyle{ruled}
\newfloat{listing}{tb}{lst}{}
\floatname{listing}{Listing}
\title{Modifying RL Policies with Imagined Actions: How Predictable Policies Can Enable Users to Perform Novel Tasks}
\author{
    Isaac Sheidlower, Reuben Aronson, Elaine Short
}
\affiliations{
    Tufts University, Department of Computer Science \\ \{isaac.sheidlower, reuben.aronson, elaine.short\}@tufts.edu
}

\usepackage{bibentry}
\usepackage{color}

 \newcommand{\algo}{IODA}
 \newcommand{\algofull}{Imaginary Out-of-Distribution Actions}

\begin{document}

\maketitle

\begin{abstract}
It is crucial that users are empowered to use the functionalities of a robot to creatively solve problems on the fly. A user who has access to a Reinforcement Learning (RL) based robot may want to use the robot's autonomy and their knowledge of its behavior to complete new tasks. One way is for the user to take control of some of the robot's action space through teleoperation while the RL policy simultaneously controls the rest. However, an out-of-the-box RL policy may not readily facilitate this. For example, a user's control may bring the robot into a failure state from the policy's perspective, causing it to act in a way the user is not familiar with, hindering the success of the user's desired task. In this work, we formalize this problem and present \algofull{}, \algo{}, an initial algorithm for addressing that problem and empowering user's to leverage their expectation of a robot's behavior to accomplish new tasks.
\end{abstract}

\section{Introduction}
As robots are increasingly in the hands of users, it is necessary to ensure that these users are empowered to use the robot and its functionality to accomplish the tasks they want. Ideally, one can deploy a robot with an exhaustive library of Reinforcement Learning (RL) policies that can autonomously perform any task the user wishes. However, this approach is impractical as users may wish to seamlessly use the robot for a task or in an environment that was not foreseen by the designers of the robot. In reality, the user will be provided with a limited set of policies and a means of controlling the robot, such as teleoperation. Given this, there is a need for methods that enable users to take advantage of these two types of control to perform new and emergent tasks.  

Consider the following illustrative example. Sally frequently uses her assistive robot arm to perform tasks around the house. Sally has used the robot arm's pick-and-place RL policy to move cups full of liquid many times. She knows how the robot will pick up the cup and steadily move it to a specified location. Now Sally wants to use the same functionality to water her flowers. To do this, she starts the autonomous policy and during execution rotates the robot's wrist to pour the water over her basin and water her flowers as the robot moves along it's path. Although in this example the interaction led to the desired result, an RL policy ``as-is'' may not facilitate this. For instance, if the RL policy was trained on a reward function such that if it spills liquid, it should stop moving along its path until the cup is upright. In this case, Sally may accidentally pour all her water onto one flower, since the robot did not continue along its path as she had expected. Similar problems exist in the autonomous driving domain, such as driving intentionally on the sidewalk or hitting an obstacle. 

These examples demonstrate potential problems of naively using an RL policy while the user controls part of the robots behavior. Since the user may have only seen successful examples of task execution, they may not know that their control signal can cause failure with respect to the robot's reward function, causing it to behave unexpectedly. Furthermore, that unexpected behavior may impede the user's ability to perform novel tasks or result in task failure from the user's perspective. This problem setting also allows the user to bring the policy ``out-of-distribution,'' which may result in undesirable behavior. There is a need to both better understand this type of interaction from a user's point of view and develop algorithms that act on RL policies to better facilitate the user completing their task. Our key insight is that whilst the user has partial control, the robot should act in a way similar to the behavior the user has seen before and is familiar with, allowing the user to leverage their prior knowledge about the robot's behavior to adapt it to new tasks.

In this paper, we formalize the aforementioned problem setting and present \algofull{} (\algo{}), an algorithm that modifies the state passed into an RL policy by projecting its current state to a previous state the user has seen: despite the robot's current real state, it chooses actions based on an \emph{imagined} state. \algo{} only modifies the state when the user brings the state out-of-distribution (OOD) with respect to states the user is familiar with/has seen before. IODA uses an OOD detector to determine when to modify the state, and then a projection function to determine the nearest state the user is familiar with to project on to. We study \algo{} in simulation and then discuss our future work studying how \algo{} can facilitate real users to accomplish new tasks. 

\section{Related Work}
Reinforcement learning (RL) is a common and effective way for robots to learn new tasks \cite{kaelbling_reinforcement_1996, kober_reinforcement_2014, ibarz_how_2021}. Much of the success of RL for robotics has come from Deep RL \cite{arulkumaran_brief_2017, nguyen_review_2019} and human-in-the-loop learning \cite{christiano_deep_2017, arzate_cruz_survey_2020, schrum_mind_2022}. However, these methods either assume that the task the user wishes to accomplish is known in advance or that a user is willing to spend time teaching a robot. These assumptions may limit the scope or applicability of any RL-based robot to one or a few tasks. While methods such as multitask reinforcement learning \cite{xie_lifelong_2022, shridhar_perceiver-actor_2023, walke_dont_2023}, large-language model skill grounding \cite{ichter_as_2023, driess_palm-e_2023, liang_code_2023}, and behavioral-diversity learning \cite{eysenbach_diversity_2018, osa_discovering_2022} have all partially addressed this problem by making robot's task repertoire larger and more diverse, they do not explicitly empower users with a method of controlling or collaborating with the robot other than through instruction.

In contrast to RL, teleoperation gives users complete control over a robot's behavior. However, the complete teleoperation of a complex robotic system can be challenging for users as there can be many degrees of freedom to control at once with limited feedback from the robot \cite{darvish_teleoperation_2023, moniruzzaman_teleoperation_2022}. A common approach to alleviate some of the burdens of teleoperation while maintaining a user's control-based empowerment is Shared Control (SC) \cite{li_continuous_2015, matsumoto_shared_2022, bustamante_toward_2021, dragan_policy-blending_2013}. SC often manifests itself as a blending of the user's and robot's control. We examine a case where the user has complete/teleoperation control over some parts of the robot while the others are fully autonomous. This is not unlike Shared Autonomy (SA) \cite{javdani_shared_2015, gopinath_human---loop_2017, selvaggio_autonomy_2021, fontaine_differentiable_2021}, where a user's control signal is interpreted as an indication of their desired goal. In SA, a typical assumption is that a user's goal is known in advance and can be represented in clear contrast to other goals in the environment. Although this assumption has been relaxed in some previous work by updating the potential set of goals \cite{zurek_situational_2021, fontaine_differentiable_2021, yoneda_noise_2023}, these approaches require a user to first demonstrate the task or a similar task without assistance. Instead, we focus on a user leveraging their creative problem-solving skills to partially control an otherwise fully autonomous robot to complete a novel task. In other words, the robot is not assisting or augmenting the user's control signal; rather, the user wants to make use of the robot's behavior as they understand it. Because of this, a robot's behavior should remain predictable under the partial control of the user. 

Legible robot motion can be defined as robot behavior that is easy for a human to predict and understand. A common way to generate legible motion in a goal-based robotic tasks is to model the user as having an internal cost function that is minimized when the robot's behavior saliently moves towards a given goal \cite{dragan_generating_2013, dragan_legibility_2013, faria_understanding_2021}. Another approach is to learn from humans through demonstrations or through feedback \cite{busch_learning_2017, bied_integrating_2020}. An important part of each of these works is that legibility and predictable robot behavior are in the context of the robot \emph{completing the task} and are often in real-time as opposed to pre-hoc or post-hoc explanations \cite{das_explainable_2021, cruz_explainable_2021, sakai_explainable_2022, paleja_utility_2021}. Predictability is also an important part of our work. We make and leverage the assumption that a robot's behavior is predictable if the user has seen similar behavior before. This assumption is not entirely dissimilar from robot-centric notions of out-of-distribution (OOD) states and behavior. 

OOD detection is an important problem in many robotic and machine learning tasks \cite{yang_generalized_2022, sun_out--distribution_2022}. Detecting when a scenario is OOD with respect to a robot's training data or what it has previously encountered may have implications about the robot's environment or performance, such as when an RL agent may behave sporadically or in an unexpected way when in a state it has never been in before \cite{lan_can_2023-1}. It has also been used to detect when a robot may require feedback from a person to help complete a novel task \cite{dass_pato_2023}. OOD in RL has recently been studied to detect when an agent is acting in a new MDP \cite{haider_out--distribution_2023}. We apply similar techniques to detect when the robot is in part of the environment that they would not otherwise act in. This can happen when a user is partially controlling a robot to perform a new task.

\section{Problem Setting}
Here we describe a problem setting in which a user is familiar with the autonomous execution of a task and wishes to partially control a robot during that execution to accomplish another task. In short, the user creates a plan to partially control the robot based on how they expect the robot to behave to accomplish a novel task. Thus, it may be critical that the robot behaves in a way that is predictable to the user, no matter where the robot may be in the state space. Given that the robot behaves in a user-predictable way, as opposed to sporadically or in an unfamiliar way, a user can perform various new tasks with relative ease and few surprises. 

In this setting, the user has seen the robot complete its task many times. We refer to this as a history of task ``rollouts.`` Based on this, we assume the following: when the robot is in a state the user has never seen before, the user expects that the robot will do the same thing as it would do in the ``closest`` state to its current state. Where ``closest`` is both problem and user specific, however, the intuition is that the robot will behave in a similar way in similar circumstances and that in novel circumstances, a user will project on to what they have seen before. This assumption temporarily constrains the problem space; however, it is a reasonable assumption in many robotic tasks. Thus, the problem can be defined as, for any given state unseen to the user, the robot should find a state that the user has seen before and act as if it were in that state.   

We will now define the original task the robot can complete autonomously and how this task is used to build up a user's expectation of the robot's behavior. Let task $orig$ be defined as an MDP with states $S\subset \mathbb{R}^n$, actions $A$, reward function $r: S \times A \rightarrow \mathbb{R}$, and transition function $T: S \times A \rightarrow S$, and discount factor $\gamma$. And there is a robot that has learned an optimal policy for the task denoted $\pi^*$. Let $D$ be defined as a history of rollouts under $\pi_{orig}^*$ that the user has seen. Then, let the user expectation of the robot's behavior given $D$ be:
\vspace{-1mm}
$$W_{D}:S \rightarrow S$$

Here, $W$ is a function that maps from the robot's current state to their anticipation of what the next state will be.

To this setting, we introduce partitioned control, where the user teleoperates one or more parts of the robot's actions. We separate the action space into two separate sets $A_U, A_R \subset A$; $A_U$ denotes the actions that the user can take and $A_R$ denotes the actions that the robot can take. We further assume that the user and robot action spaces are \emph{disjoint}; that is, $A_U \cap A_R = \{\mathbf{0}\}$. In other words, the user and the robot control different parts/axes of the action space. For example, if the robot is acting in Cartesian space, the user may take control over x-axis actions, or take control over the rotation of a specific joint. We denote the user's expectation of how the robot will act with their partial control signal as:
\vspace{-.9mm}
$$W_{D, U}:S \rightarrow S$$

where $U$ is the user control. For brevity, hereafter we refer to this only as $W.$ 

In order to make the robot behavior more predictable for the user, we want to adjust the behavior of the robot policy when it is outside the user's observation set $D$. The goal is to identify when the robot is in a novel state $s$ where there exists a state $s' \in D$ that leads to more predictable behavior. Formally, identify when $\exists s' \in D$ s.t.:
\vspace{-.9mm}
 \begin{equation}
     d(W(s), T(s, u \circ \pi^*(s))) \geq d(W(s), T(s, u \circ \pi^*(s')))
 \end{equation}

 where $d$ is a task-dependent distance metric between states, and $u\circ\pi^*(s)$ denotes the disjoint combination of the autonomous action of the robot and the user's teleoperation. When such a $s'$ is identified, the robot should act as if it were in $s'$. Specifically, we want to select a new proxy state $s'$ such that the user's \emph{predicted} state $W(s)$ is closer to the actual resultant state when simulating the policy in $s'$, $T(s, u \circ \pi^*(s'))$  than to the resultant state of running the policy directly $T(s, u \circ \pi^*(s))$. Lastly, in this setting, the true $W$ and the nature of the new task that the user wishes to accomplish are unknown. However, formalizing $W$ as such can be useful for modeling and/or simulating, creating a learning objective, or creating metrics to measure the success of algorithms applied to this problem. 

\section{\algofull{} (\algo{})}
In this section, we present \algofull{} (\algo{}), to facilitate a user to accomplish new tasks given a policy and a means of teleoperating an axis of robot behavior (the setting described in the previous section). Our key insight is that when the robot is acting in a region that greatly differs from what the user has seen before, the policy should act with imagined states that are as similar to the real state as possible while being ``in-distribution'' of what the user is familiar with and anticipates. For the rest of this section, unless otherwise specified, we refer to ``in/out of distribution'' states with respect to $D$. 

\begin{algorithm}[t]
\caption{\algofull{} \\ (\algo{})}
\label{alg:algorithm}
\textbf{Initialize:} Rollout history D \\
\textbf{Initialize:} OOD-state detector \\
\textbf{Initialize:} State $s$
 
\begin{algorithmic}[1] 
\WHILE{not done}
\IF {s is OOD}
\STATE $s' \rightarrow \arg\min_{s'\in D}d(s,s')$ 
\STATE $a \rightarrow \pi^*(s')$
\ELSE 
\STATE $a \rightarrow \pi^*(s)$
\ENDIF
\STATE $u \rightarrow $ user's control signal
\STATE $s_{t+1} \rightarrow T(s, u\circ a)$
\ENDWHILE
\end{algorithmic}
\end{algorithm}

The complete \algo{} algorithm is presented in Algorithm 1. Here, we require that an OOD detector be trained on $D$. This is then used to detect ``novel'' states. While this technique is not new from a robot-centered perspective, it is also a human-modeling choice that draws an analogy between when an OOD detector outputs a state that is OOD and when a human may be projecting to a state they have seen in the past. It is also being used to determine when to search for a state that the robot policy should ``imagine'' it is in.

\begin{figure*}[t]
  \centering
  \subfloat[IODA]{\includegraphics[width=0.33\textwidth]{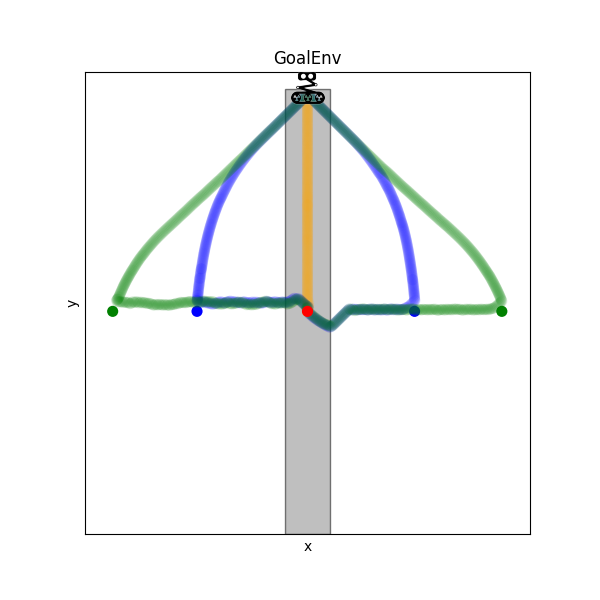}\label{fig:f1}} 
  \hfill
  \subfloat[Out-of-distribution, \\ no imagined state]{\includegraphics[width=0.33\textwidth]{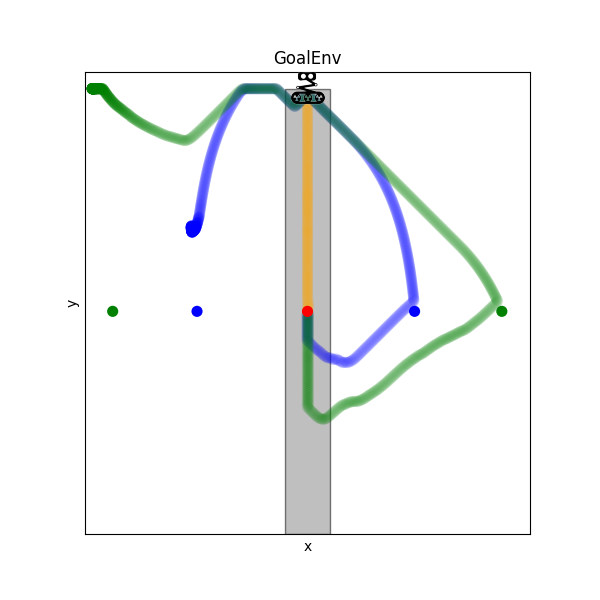}\label{fig:f2}}
  \hfill
  \subfloat[Must return to workspace, \\ no imagined state]{\includegraphics[width=0.33\textwidth]{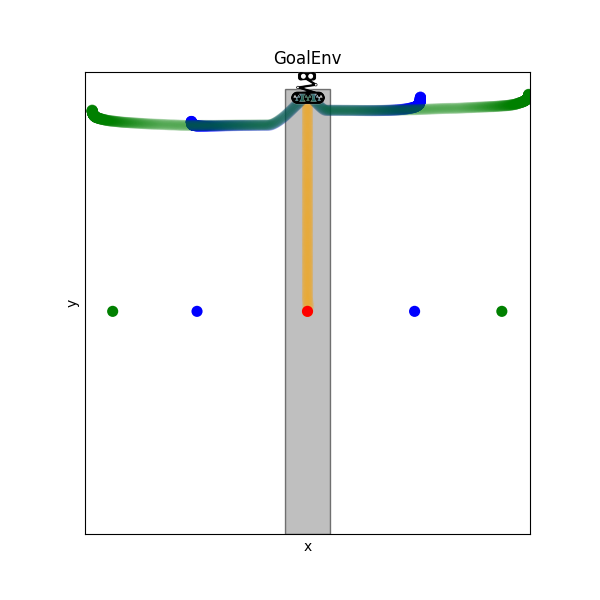}\label{fig:f3}}
  \vspace{-.8mm}
  \caption{In a 2D goal navigation environment, a simulated user is trying to leverage an optimal policy to reach subgoals by controlling the x-axis of the robot whilst the policy controls the y-axis. These subgoals are outside the robot's workspace (highlighted in gray). Our algorithm \algo{} allows the user to seamlessly reach the subgoals.}
  \label{2dsim}
  \vspace{-4mm}
\end{figure*}

\section{Simulation Example}
In this section we demonstrate in a 2d continuous navigation task both the problem setting and the application of the \algo{} algorithm. In the original task (see Figure 1), the robot learned to navigate to any specified goal point from within its workspace (highlighted in gray in Figure 1). The reward function for the task is defined as $-|p_{agent}-p_{goal}|$, or the negative distance of the agent's position to the goal. The user then wishes to leverage the optimal behavior they have observed to control the x-axis actions to first guide the robot to an intermediate goal outside of the workspace (shown in Figure 1 as blue or green dots) and then to the primary goal: a novel task not represented by the robot's policy.

We train two RL agents to optimally solve two slightly different versions of the original task using the off-policy RL algorithm SAC \cite{haarnoja_soft_2019}. We use SAC as it has been shown to be relatively robust when acting out-of-distribution \cite{lan_can_2023-1}. In one version of the task, Figure 1, b, the RL algorithm was trained constrained to the gray workspace, simply being penalized for leaving it. In the other version, Figure 1, c, the agent is penalized if it is out of the workspace, and is further penalized by moving in the y-axis whilst it is out of the workspace. This encourages the agent to return to its workspace as quickly as possible before continuing the task. In both versions, the agent, when out of its workspace,  may behave in a way unpredictable to a user. As a user has only ever seen optimal rollouts, they may not be familiar with what happens when the robot ``fails`` or is out-of-distribution.

In these environments, we collected 1000 rollouts of the optimal policies and trained Deep SVDD OOD detectors \cite{ruff_deep_2018} on the states of those rollouts. We choose $d$ to be the $L1$ distance between two states. Lastly, we substitute a human's control for an optimal x-position controller given the current x-position and subgoal location. As can be seen in Figure 1. \algo{} is the only condition in which the simulated user can reach all subgoals and then easily go to the primary goal. In Figure 1, b, the agent acted relatively sporadically when brought out-of-distribution, and could only reach both goals half the time.  In Figure 1, c, since the agent was trained not to move in the y-axis when outside of its workspace, the policy of the simulated user was incompatible with the agent's behavior. Furthermore, $D$ did not contain any indication that the robot would stop.

\section{Future Work and Discussion}
It is crucial that users are empowered to use the functionalities of a robot to solve creative and on-the-fly problems. We believe this can be achieved without necessarily modeling a user's internal goals or trying to account for everything a user may want in advance but rather by providing second-order tools that a user can use to adjust and freely modify agent behavior. In the case of reinforcement learning, policies often have failure cases or behavior that may not be seen by a user during day-to-day use cases, and this can inhibit a user's creative use of a policy to perform new tasks without having to explicitly teach the robot. By keeping track of not just the robot's experience but also the user's experience with the robot, we can reason about how to facilitate a user to leverage robot behavior they have seen before.

In this work, we presented a proof-of-concept problem setting and the \algo{} algorithm. This problem setting is important to examine and study to ensure end-users are empowered to perform novel tasks. In our follow-up work, we plan on expanding this setting. For example, in this work, we study the problem setting in a task where the notion of distances between states is fairly well defined and an agent's actions in those states are fairly predictable. Furthermore, we plan to empirically investigate our assumptions about how humans perceive an agent both when they partially control them and when acting independently. This investigation will be done through a user study where users can indicate predictions about an agent's behavior given what they have seen before. We then plan on a follow-up user study in significantly more complicated tasks than in the simulation environment, such as plant watering and kitchen organizing. The second study will examine a user's ability to perform these novel tasks given an RL policy and a means of teleoperation through \algo{} and other baselines. 

\section{Conclusion}
We proposed that users can simultaneously use partial teleoperation in conjunction with an RL policy to perform novel tasks. This is based on the claim that users anticipate a robot's behavior under their partial control given that they have seen and are familiar with how it behaves autonomously. Unfortunately, a robot's behavior may not always meet a user's expectation. Rather, when a user takes partial control, they may unintentionally bring the RL policy into failure states or ``out-of-distribution,'' both of which may cause unexpected robot behavior and hinder the success of the desired task. To address this, we proposed the \algo{} algorithm, which passes imaginary states to the RL policy that result in more predictable behavior. We demonstrated \algo{} in a simulation and plan to study both the problem setting and \algo{} in the hands of real users.

\bibliography{main}
\end{document}